\documentstyle[open,psfig] {icslp98}

\procyear{98}
\title{A Computational Memory and Processing Model for Prosody}

\procyear{98}
\author{Janet E. Cahn}
\affiliation{Massachusetts Institute of Technology \\
cahn@media.mit.edu}

\begin{document}

\maketitle

\begin{abstract}

{This paper links prosody to the information in the text and how it is processed by
the speaker. It describes the operation and output of {\sc Loq}, a text-to-speech
implementation that includes a model of limited attention and working memory.
Attentional limitations are key. Varying the attentional parameter in the
simulations varies in turn what counts as given and new in a text, and therefore,
the intonational contours with which it is uttered. Currently, the system produces
prosody in three different styles: child-like, adult expressive, and knowledgeable.
This prosody also exhibits differences within each style -- no two simulations are
alike. The limited resource approach captures some of the stylistic and individual
variety found in natural prosody.} 

\end{abstract}

\section{Introduction}
 
Ask any lay person to imitate computer speech and you will be treated to an
utterance delivered in melodic and rhythmic monotone, possibly accompanied by choppy
articulation and a voice quality that is nasal and strained. In fact, current
synthesized speech is far superior. Yet few would argue that synthetic and natural
speech are indistinguishable. The difference, as popular impression suggests, is the
relative lack of interesting and natural variability in the synthetic version. It
may be traced in part to the lack of a common causal account of pitch, timing,
articulation and voice quality. Intonation and stress are usually linked to the
linguistic and information structure of text. Features such as pause location and
word duration are linked mainly to the speaker's cognitive and expressive
capacities, and pitch range, intensity, voice quality and articulation to her
physiological and affective state. 

%ladd:etal85}.

In this paper, I describe a production model that attributes pitch and timing to the
essential operations of a speaker's working memory -- the storage and retrieval of
information. Simulations with this model produce synthetic speech in three of the
prosodic styles likely to be associated with attentional and memory differences: a
child-like exaggerated prosody for limited recall; a more adult but still expressive
style for mid-range capacities; and a knowledgeable style for maximum recall. The
same model also produces individual differences within each style, owing to its
stochastic storage algorithm. The ability to produce both individual and genre
variations supports its eventual use in prosthetic, entertainment and information
applications, especially in the production of reading materials for the blind and
the use of computer-based autonomous and communicative agents. 

\section{A Memory Model for Prosody}

Prosody organizes spoken text into phrases, and highlights its most salient
components with {\em pitch accents}, distinctive pitch contours applied to the
word. Pitch accents are both attentional and propositional. Their very use indicates
salience; their particular form conveys a proposition about the words they mark. For
example, speakers typically use a high pitch accent (denoted as H*) to mark salient
information that they believe to be {\it new} to the addressee. Conversely, when
they believe the addressee is already aware of the information, they will typically
de-accent it\cite{brown83} or, if it is salient, apply a low pitch accent
(L*)\cite{pierrehumbert:hirschberg90}. Re-stated as a commentary on working memory,
the H* accent conveys the speaker's belief that the addressee can not retrieve the
accented information from working memory. De-accenting implicitly conveys the
opposite expectation. The L* accent does so explicitly. This view predicts different
speaking styles as a consequence of the speaker's beliefs about an addressee's
storage and retrieval capacities. For example, it ascribes the exaggerated
intonation that adults use with infants and young children\cite{fernald:simon84}, to
the adults' belief that the child's knowledge and attention are extremely limited;
therefore, he needs clear and explicit prosodic instructions as to how to process
language and interaction.

The model of working memory I use shows how retrieval limits can determine
the information status of an item as either given or new, and therefore,
its corresponding prosody. It was developed and implemented by Thomas
Landauer\cite{landauer75} and models {\em working memory} as a periodic
three dimensional Cartesian space, the {\em focus of attention} via a
moving search and storage {\em pointer} that traverses the space in a slow
random walk, and {\em retrieval ability} via a {\em search radius} that
defines the size of a region whose center is the pointer's current
location. Search for familiar items proceeds outward from the pointer, one
city block per time step, up to the distance specified by the search
radius.

As a consequence of the random walk, incoming stimuli are stored in a
spatial pattern that is locally random but globally coherent. That is,
temporal proximity in the stimuli begets spatial proximity in the model. It
contrasts with stack models of memory that are strictly chronological, and
semantic spaces in which distance is conceptual rather than temporal. Most
importantly, it is a valid computational model of attention and working
memory (AWM, from here on). Landauer used it to reproduce the well-known
learning phenomena of recency and frequency, in which subjects tend to
recall stimuli encountered most recently or most
frequently\cite{landauer75}. It has since been used by
Walker\cite{walker95} to show that resource-bound dialog partners will make
a proposition explicit when it is not retrievable or inferable, despite
having been previously mentioned.

\begin{figure}
\centerline{\psfig{figure=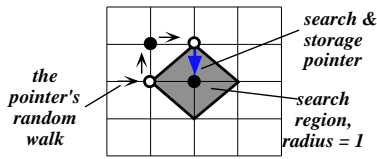,width=50truemm}}

\caption{Using AWM, stimuli are classed as given if they have counterparts within
the the search radius. New items have no such counterparts because they are either
not in working memory, or are stored outside the radius.} 

\label{fig--awm-ex} 
\end{figure}

Retrieval in AWM is the process of {\em matching} the current stimulus to
the contents of the region centered around the pointer. The search radius
determines the size of this region and therefore is the main AWM simulation
parameter. If a
match is found within the search region, the stimulus is classified as
given, otherwise, it is new. Figure~\ref{fig--awm-ex} illustrates this with
the simple example of filled and unfilled circles, a 4x4 AWM space, and a
search radius of one. At the center of the search region is the current
stimulus, a filled circle. Because the region contains no other filled
circles, the stimulus is classed as new. Had the  stimulus been an
unfilled circle, it would have instead been classed as given because a
match is retrievable within the search radius. Or, alternatively, had the
search radius been two instead of one, a matching filled circle would have
been found, and the stimulus again classed as given.

The ability to identify given and new items makes AWM a useful producer of prosody
based on this distinction. Ostensibly, it shows how a speaker's processing affects
her prosody. However, although the working memory belongs to the speaker, its
operation and determinations may reflect the speaker's own retrieval capacities, her
estimate of those of the addressee, or a mixture of both. That is, a speaker can
always adapt her style (prosodic and lexical) to the needs of a less knowledgeable
or capable addressee. A cooperative and communicative speaker will usually do this.
However, she cannot model a retrieval capacity greater than her own -- her own
knowledge and attentional limits always constitute the upper bounds on her
performance. 

\section{System design}

\begin{figure}
\centerline{\psfig{figure=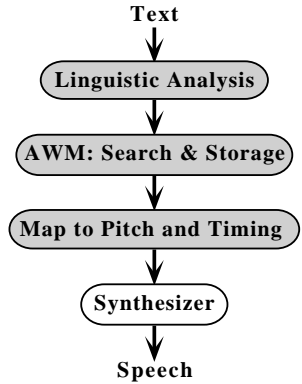,width=40truemm}}
\caption{AWM as a component of the {\sc Loq} system.}
\label{fig--model-vert} 
\end{figure}

The AWM component is embedded in a software implementation, {\sc Loq}, that takes a
text-to-speech approach. As shown Figure~\ref{fig--model-vert}, the input to AWM is
text, the output is speech. Therefore, {\sc Loq} models read rather than spontaneous
speech. Text comprehension is the process of searching for a match. Uttering the
text is a question of mapping the search process and its results to prosodic
features and sending the prosodically annotated text to the synthesizer. 

Like many commercial text-to-speech synthesizers, the text structure is
analyzed before prosody is assigned. However, the {\sc Loq} analysis is
richer. It takes advantage of on-line linguistic databases to approximate
the speaker's knowledge of English semantics, pronunciation and usage. The
structural analysis is richer as well, providing both grammatical structure
(subject, verb, object), empty categories (ellipses, for example) and
information about clausal attachment. The main qualitative difference is
that {\sc Loq} interposes a model of limited attention and working memory
between the text analysis and prosodic mapping components.

\subsection{Matching}

For the example in Figure~\ref{fig--awm-ex}, the matching criterion is binary and
simple -- a circle is either filled or unfilled. However, language is many more
times complex, and matches may occur for a variety of features, some of which are
more informative than others. The matching criteria used in {\sc Loq} attempt to
distill from the literature (e.g., \cite{nooteboom:terken82,fay:cutler77}) the most
relevant and prevalent ways that items in memory {\em prime} for the current
stimulus, and by the same token, the ways in which the current stimulus can function
as a {\em retrieval cue}. In other words, they gauge the mutual information between
the current stimulus and previously stored items.

\begin{figure}
\centerline{\psfig{figure=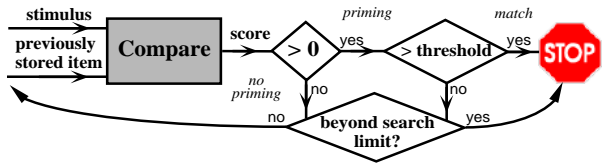,width=80truemm}}
\caption{{\sc Loq} matching algorithm.}
\label{fig--matching} 
\end{figure}

Altogether, {\sc Loq} tests for matches on twenty-four semantic, syntactic,
collocation, grammatical and acoustical features. Each test contributes to the total
match score, which is then compared to a {\em threshold}. If it is below, the search
continues; if above, it stops. As shown in Figure~\ref{fig--matching}, matches on
any criterion express priming, and scores above the threshold constitute a match
sufficient to stop the search even before it reaches the edge of the search region.
Because some tests are more informative than others, a high score can reflect the
positive outcome of many un-informative tests, or of one that is definitive. Thus,
in the current ordering, co-reference ensures a match, while structural parallelism
in and of itself does not.

\subsection{Input}

The matching criteria determine the form and kind of information in the text input.
As with commercial synthesizers, this includes part of speech tagging. {\sc Loq}
uses the output of Lingsoft's ENGCG (English Constraint Grammar) software which
provides both tags and phrase structure information. However, reliable automatic
means for identifying other information, such as grammatical clauses, empty
categories, attachment and co-reference do not yet exist. Therefore, this
information was entered by hand.

The {\sc Loq} software turns the parsed and annotated text into a sequence of
tokens that assembles clauses in a bottom up fashion, starting with the
word and followed by the syntactic and grammatical clauses to which it
belongs. This models the reader's assembly of the words into meaningful
syntactic and grammatical groupings.\footnote{Adapting this for a
spontaneous speaker would proceed in reverse, from the concept, to
grammatical roles, syntactic phrases and finally, the words.}

To facilitate the matching process, the text is also augmented with information from
the WordNet database for semantic comparisons, a pronunciation
database for acoustical comparisons and the Thorndike-Lorge and Kucera-Francis for
word frequency counts\footnote{As provided in the Oxford Psycholinguistic
Database.} to scale the match score by the prior probability for the language. The
WordNet synonym indices were assigned by hand. However, all subsequent semantic
comparisons using WordNet are automatic as required by the matching process.

\subsection{Mapping}

I have described how AWM produces the L* accent (or none) for retrievable items, and
H* for new ones. However, there are more than two pitch accents -- Pierrehumbert
{\it et al.}\cite{beckman:pierrehumbert86} identify six\footnote{L*, H*, L+H*, L*+H,
H+L*, H*+L.} -- and more components to prosody. Obtaining them from one model first
requires an adjustment such that given or new status is determined from the effect
of the stimulus on the region as a whole, as follows: The result of any one
comparison affects the ``state'' of the item to which the stimulus is compared.
State is simply defined -- a L annotation records a match most any
criterion,\footnote{Some criteria are parasitic and only contribute to the score in
combination with other criteria.} and a H annotation records a match score of zero.
Thus, the comparison process registers both priming and a true match. Both receive L
annotations, but only a match whose score exceeds the threshold stops the search. 

A pitch accent is then derived by comparing the contents of the search radius {\it
before} and {\it after} the matching process. Majority rules apply such that the
annotation with the higher count becomes the defining tone. If both the before and
after configurations are composed mainly of L annotations, the accent form is L+L,
which becomes the L* accent. However, if there is a change, for example, from a L to
H majority, the accent form is L+H. The interpretation of L+L is, roughly, that a
familiar item was expected and provided. Likewise, the interpretation for L+H is
that a familiar item was expected but an unexpected one provided.

To complete the bitonal derivation, {\sc Loq} treats the location of the main tone
as a categorical reflection of the magnitude of the effect of the stimulus. If the
stimulus changes the annotations for the majority of items in the search region, the
second tone is the main tone. Otherwise, it is the first. This schema produces the
six pitch accents identified by Pierrehumbert {\it et al.} More generally, the
annotation schema provides the model with a simple form of feedback -- the results
of prior processing persist and contribute to a bias that affects future processing.

\begin{figure}
\centerline{\psfig{figure=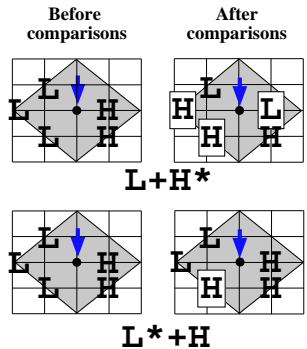,width=40truemm}}
\caption{In {\sc Loq}, bitonals occur when the L/H counts differ before and after
the matching process. The main tone of a bitonal is treated as a categorical
indicator of the magnitude of the effect of the stimulus on the context.}
\label{fig--bitonals} 
\end{figure}

The pitch accent mapping illustrates the main features of the prosodic mapping in
general. First, all mappings reflect the activity and state within the region
defined by the search radius. Second, they express some aspects of prosody as a
plausible consequence of search and storage. For example, storage and search times
are mapped to word and pause duration. However, others -- for example, the bitonal
derivation-- are, at best, coherent with the operation and purpose of the model and
not contradicted by the current (sparse) data on the relation of cognitive capacity
to the prosody of read speech. In all, the mapping from AWM activity and state
produces intonational categories (pitch accent, phrase accent and boundary tone)
and their prominence, word duration, pause duration and the pitch range of an
intonational phrase. 

\section{Results}

Although simulations were run using text from three different genres (fiction, radio
broadcast, rhymed poetry), two and three dimensional AWM spaces and three memory
sizes (small, mid-range and large), most of the prosodic output was correlated with
the search radius. Therefore, the results reported here are for the mid-range
two-dimensional memory (22x22) and for the news report text only (one paragraph, 68
words.) Five simulations were run for each radius.

\begin{figure}
\centerline{\psfig{figure=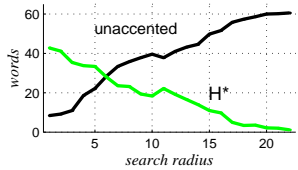,width=40truemm}}

\caption{Mean unaccented {\it vs.} H* accented word counts as a function of search
radius.} 

\label{fig--accents} 
\end{figure}

\begin{figure}
\centerline{\psfig{figure=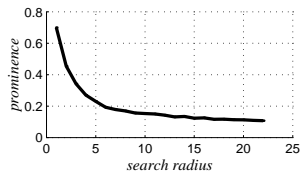,width=40truemm}}
\caption{Mean pitch accent prominence for all six accent types, as a function of search 
radius. } 
\label{fig--pa-prom} 
\end{figure}

True to the attentional predictions, Figure~\ref{fig--accents} shows that as the
search radius increases, the mean number of unaccented words increases as well,
while the number of H* accents decreases. Under the current mapping, pitch accent
prominence is a function the distance at which the search stops and the number of
comparisons performed prior to stopping. This produces a decrease in the mean
prominence as the search radius increases (Figure~\ref{fig--pa-prom}). These patterns
contribute to the lively and child-like intonation produced for the smallest radii
(1 and 2), the expressive but more subdued intonation for the mid-range radii (3-8)
and flatter intonation of the higher radii. 

The naturalness of synthetic prosody is difficult to evaluate via in perceptual
tests\cite{ross:ostendorf96}. However, informal comments from listeners revealed
that while the three styles were recognizable and the prosody more natural-sounding
than the commercial default, it was best for shorter sections rather than for the
passages as a whole. A comparison with the natural prosody for the same text (the BU
Corpus radio newscasts) showed that when the simulations agreed on pitch accent
location and type, they tended to disagree on boundary location and type, mostly
because the {\sc Loq} simulations produced many more phrase breaks than the natural
speaker. 

\section{Conclusion and Future Directions}

{\sc Loq} is a production model. It produces prosody as the consequence of cognitive
processing as modeled by the AWM component. Its focus on retrieval makes it a
performance model as well, demonstrating that prosody is not determined solely by
the text. It produces three recognizable styles that appear to correlate with
retrieval capacities as defined by the search radius: child-like (for radii of 1
and2), adult expressive (for radii between 3 and 8) and knowledgeable (for radii
higher than 8). This is a step towards producing prosody that is both expressive and
natural and, in addition, specific to the speaker,

Currently, the main problem is that the prosody is not entirely cohesive within one
text. Therefore, one next step is to explore variations on the mapping of AWM
activity and state to prosodic features. More distant work includes extending the
model to incorporate other influences, especially the influence of physiology. This
may be the key to producing more than three styles, and to incorporating both the
dynamics and constraints that will produce consistently natural-sounding speech.

%\bibliography{/u/cahn/bib/bib}
\end{document}